\documentclass[runningheads]{llncs}
\usepackage{amsmath}
\usepackage{graphicx}
\usepackage{hyperref}
\usepackage{float}
\usepackage{subcaption}

\begin{document}
\title{Towards Creating a Deployable Grasp Type Probability Estimator for a Prosthetic Hand}
\titlerunning{Deployable Grasp Estimator for a Prosthetic Hand}
%

\author{Mehrshad Zandigohar \and Mo Han \and Deniz Erdo{\u{g}}mu\c{s} \and Gunar Schirner}
\authorrunning{M. Zandigohar et al.}
%
\institute{Northeastern University, Boston MA 02115, USA \\
\email{\{zandi,han,erdogmus,schirner\}@ece.neu.edu}}

\maketitle              
\begin{abstract}
For lower arm amputees, prosthetic hands promise to restore most of physical interaction capabilities. 
This requires to accurately predict hand gestures capable of grabbing varying objects and execute them timely as intended by the user. 
Current approaches often rely on physiological signal inputs such as Electromyography (EMG) signal from residual limb muscles to infer the intended motion. However, limited signal quality, user diversity and high variability adversely affect the system robustness.
Instead of solely relying on EMG signals, our work enables augmenting EMG intent inference with physical state probability through machine learning and computer vision method.
To this end, we: (1) study state-of-the-art deep neural network architectures to select a performant source of knowledge transfer for the prosthetic hand, (2) use a dataset containing object images and probability distribution of grasp types as a new form of labeling where instead of using absolute values of zero and one as the conventional classification labels, our labels are a set of probabilities whose sum is 1. The proposed method generates probabilistic predictions which could be fused with EMG prediction of probabilities over grasps by using the visual information from the palm camera of a prosthetic hand. 
Our results demonstrate that InceptionV3 achieves highest accuracy with 0.95 angular similarity followed by 1.4 MobileNetV2 with 0.93 at {\raise.17ex\hbox{$\scriptstyle\sim$}} 20\% the amount of operations.

\end{abstract}

\section{Introduction}

Prosthetic hands aim to compensate part of the lost ability of lower arm amputees. In order to correctly enact the intent of the user, prosthetic hands consider individual finger motion  control, grasp type selection, and open close commands. In this work we focus on grasp type selection.

State-of-the-art approaches try to classify the amputee's Electromyography (EMG) signals of the residual limb muscles into meaningful motions. This approach has drawbacks which adversely affect its robustness in real life situations \cite{EMG,EMG_CSL}. For instance, they need calibration pretty often; the unexpected electrode shifting could distort the EMG signals; muscle fatigue and/or limb disposition adversely affect the EMG patterns; and some amputees may lack critical muscles which EMG classification rely on. These insufficiencies have led researchers to use more sources of information to understand human intent \cite{RANKCNN}. With the rise of Convolutional Neural Networks \cite{CNN,video2019,VGG16,imagenet,googlenet,inceptionV3,resnet50}, studies on using visual information as a source of information for the prosthetic hand have been conducted \cite{2015exploratory,2016automatic,2017visual,2017deep,Depth2014microsoft,RANKCNN}, which focus on classifying images into a grasp type. 

\begin{figure}[t]
  \centering
  \includegraphics[width=\linewidth]{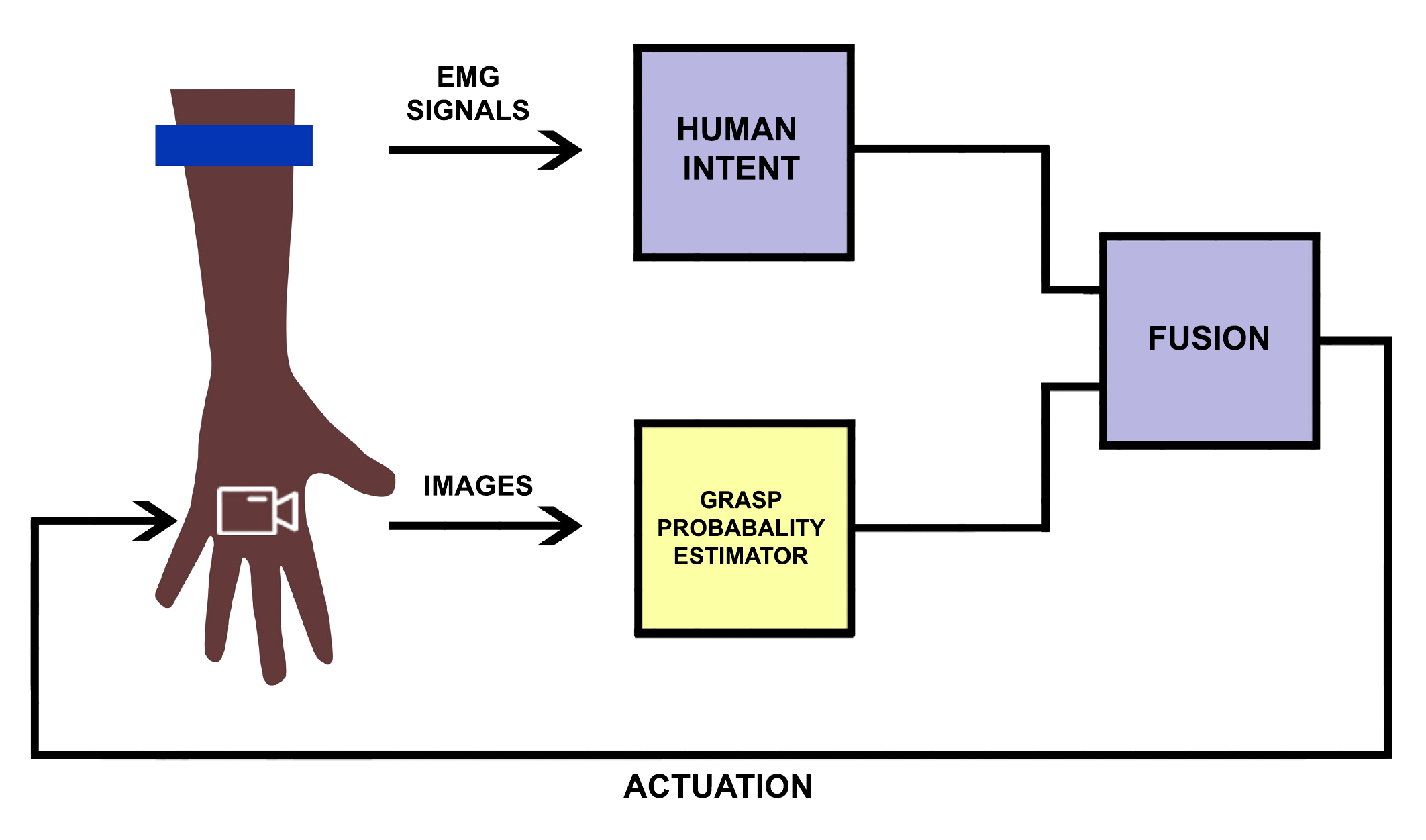}
  \caption{System components and flow of information.}
  \label{fig:system-overview}
\end{figure}

\autoref{fig:system-overview} demonstrates the overview of our prosthetic hand design, where an EMG sensor is attached to user's arm and the collected EMG signals are used to infer the human intent while the grasp probability estimator provides physical state information using the images captured from the palm camera of the prosthetic hand. The resulting predictions from both EMG and vision are then combined in Fusion module to form a final decision. This work focuses on the grasp probability estimation to enable efficient and accurate fusion of the physical information. 

The conventional vision-based classifiers try to estimate the probable grasp type based on absolute values of one and zero assigned to each class as labels during training. However, in the context of grasp estimation, this approach has 2 drawbacks: (1) not every object is limited to a single gesture capable of grabbing that object, i.e. there might be more than a way to grab an object; (2) Each potential grasp type for a specific object might not have the same level of preference for different users. Therefore, the predicted value for each grasp type could be represented much more accurately as probabilities over grasps which sum up to 1. Therefore, to facilitate physical information from the camera with human intent provided by EMG, probability based estimation yields much more information over fusing the absolute predictions. 

On the other hand, while having an accurate prediction helps with inferring the human intent for selecting a grasp type, given the real-time constraints of inference for prosthetic hands, the network should behave in a timely manner to meet real-time deadlines.  In this work we:
\begin{enumerate}
    \item Provide a probability distribution based neural network rather than training on absolute values of zero and one to capture the true nature of capable gestures for grabbing a specific object and provide more information of the possible grasp types when combined with EMG inference probabilities. 
    \item We also study performance of different networks based on the amount of computation each network has to select efficient architectures to transfer which enables embedded and real-time predictions considering their limited compute power. 
\end{enumerate}

Using the proposed method, InceptionV3 as our best probability estimator reaches 0.95 angular similarity followed by 1.4 MobileNetV2 with 0.02 loss while improving performance by 5.02X. As a result, fusion of visual data with EMG data is enabled considering performance limitations. Details on EMG and fusion modules are out of the scope of this paper.

The paper continues with \autoref{sec:prosthetic-hand} providing the overview of our prosthetic hand. Following that, \autoref{sec:efficient-transfer} our method for selecting an efficient transfer source. \autoref{sec:probability-estimation} presents the details of training. \autoref{sec:results} evaluates the proposed method and provides results, and finally \autoref{sec:conclusion} concludes this paper.

\section{Prosthetic Hand}
\label{sec:prosthetic-hand}

\begin{figure}[t]
  \centering
  \includegraphics[width=\linewidth]{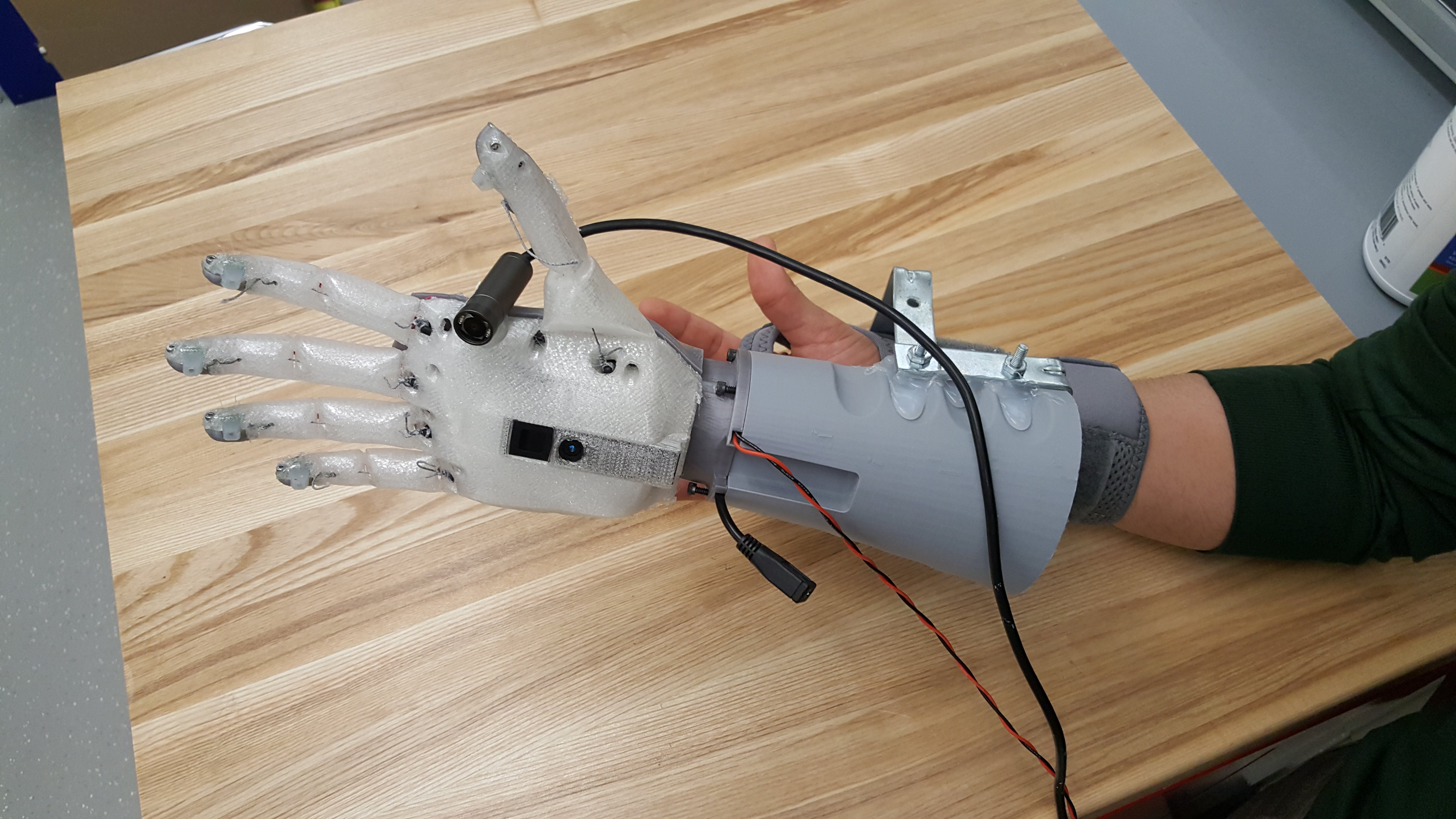}
  \caption{Prosthetic hand with camera attached.}
  \label{fig:product}
\end{figure}

\autoref{fig:product} shows the actual prosthetic robotic hand produced by our collaborators. The hand is a 3-D printed model of OpenBionics hand \cite{openbionics} with a USB endoscopic camera attached on its side. The actuators are position controlled, and would stop actuation once the drawn current exceeds some threshold. All components use ROS Melodic Morenia \cite{ros} for communicating between themselves, i.e. EMG sensor, embedded camera, actuator and fusion units. 

\autoref{fig:system-overview} demonstrates high level system integration and flow of information in this work. To predict the intent of the amputee, a classifier on the EMG signal is used in the form of a probability distribution over five grasp types. Moreover, the system is augmented with a visual grasp type probability estimator fed with the images from an embedded camera in the hand to output another distribution over the same grasp types. This is merely based on the feedback from the environment to compensate EMG deficiencies. Given the information from human intent and visual characteristics, the fusion unit aggregates the two probability distributions into the most probable grasp type considering the confidence of each unit. To have a more reliable decision, this process is repeated and averaged over two seconds to make the final decision. This decision is further sent to the control unit in order to actuate the prosthetic robotic hand to execute the grasp.

In order for the fusion unit to make the most timely decision, the visual grasp probability estimator should process as many frames as the camera generated per second to be real-time, which is 30fps in our work. As we target mobile deployment of vision system, challenges arise due to low power constraints and limited compute performance. Therefore, smaller models with less number of computations are preferred to the more accurate but computationally intensive networks. While there are researches for deployment of DNNs to an embedded target, including optimizations such as pruning, quantization, tensor fusion, kernel auto-tuning, multi-stream execution, dynamic tensor memory and precision calibration, these are outside the scope of this paper and we try to provide a platform-independant approach for selecting efficient networks.

\section{Selecting Efficient Transfer Architectures}
\label{sec:efficient-transfer}

\begin{figure}[t]
  \centering
  \includegraphics[width=\linewidth]{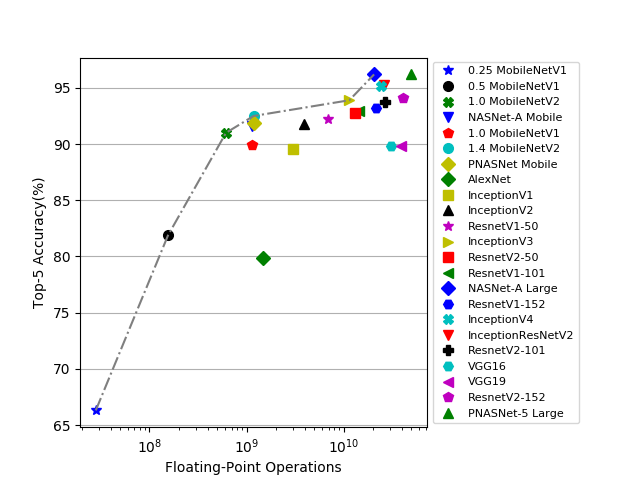}
  \caption{Accuracy vs. number of floating point operations for several pretrained ImageNet models.}
  \label{fig:pareto}
\end{figure}

While deep neural networks (DNNs) have shown promising results on many tasks, due to their tremendous number of parameters (i.e. over 25.6M parameters in ResNet50), fully training DNNs from scratch requires a large set of data. This challenge is well solved by transfer leaning. Transfer learning has shown impressive results on applications with similar domains where there is not enough data or computing resources to train a deep network from scratch \cite{transfer}. In computer vision studies, ImageNet \cite{imagenet} has been the most widely used benchmark on problems including but not limited to transfer learning \cite{imn-trans,imn-trans2}, object detection \cite{imn-det} and image segmentation \cite{imn-seg,imn-seg2}. Torralba and Efros \cite{Torralba} show that many datasets before Imagenet were biased and not general enough to be transferred to other domains. 

There are many networks trained on the large ImageNet dataset including AlexNet \cite{alexnet}, VGG \cite{VGG16}, MobileNet \cite{mobilenet}, ResNet \cite{resnet50}, Inception \cite{inceptionV3}, and NASNet \cite{nasnet} with different accuracies and architectural differences. To choose an architecture as the transfer source for the new task, one can choose the most accurate model as the base model. In \cite{google-transfer}, authors has shown that better ImageNet models provide better feature layers for transferring the learned knowledge from one domain to another. However, more accurate models generally consist of more parameters and demand more computation which results in poor inference performance.

As an initial platform-independent indicator of computation demand, this paper focuses on number of floating point operations. This allows for a high-level reasoning to compare different neural networks relatively with each other and explore the computation demand vs. accuracy trade-off. 
While suitable for this purpose, the number of floating point operations is not a substitute for estimating execution time on an actual deployment target. This would require taking into account significantly more implementation detail, such as: deployment target (CPU, GPU, neural-network accelerator), float vs. fixed point and quantization, various target-dependent optimizations. We consider as a second phase the target-dependent exploration. This work focuses on the first phase of network exploration and training. 

To this end, we have studied 23 pretrained ImageNet models given our accuracy and performance objectives depicted in \autoref{fig:pareto}. The vertical axis provides the Top-5 accuracy of each model and the horizontal axis shows the number of floating point operations required to execute inference for that network in log scale. In general, we can observe that as the accuracy of a model improves, the number of floating point operations also increase.

To avoid training all models which takes tremendous effort and time, we exploit multi-objective selection also known as Pareto Efficiency \cite{pareto} for selecting the efficient models. Given a system with function $f: R^n \rightarrow R^m$, feasible decisions $X$ is related to feasible criterion vectors $Y$ as follows:

\begin{equation}
    Y = \{ y \in R^m : y = f(x), x \in X \}
\end{equation}
and therefore the Pareto Frontier is:
\begin{equation}
    P(Y) = \{ y' \in Y : \{ y'' \succ y' , y'' \ne y' \} = \emptyset \}
\end{equation}
where the efficient models are those on the Pareto Frontier, $P(Y)$, demonstrated by the dash-dotted line in \autoref{fig:pareto}. 

Using the aforementioned method, InceptionV3, 1.4 MobileNetV2, 1.0 MobileNetV2, 0.5 MobileNetV1, 0.25 MobileNetV1 and NASNet-A Large are selected as efficient models. The selected networks strictly dominate other models and are not dominated by any other, and hence lie on Pareto Frontier. The selected networks cover the most efficient yet effective architectures. NASNet-A Large is excluded here due to very large size of the network, making training impossible on the current infrastructure.

\section{Transfer Learning}
\label{sec:probability-estimation}

\begin{figure}[t]
  \centering
  \includegraphics[width=3.5in]{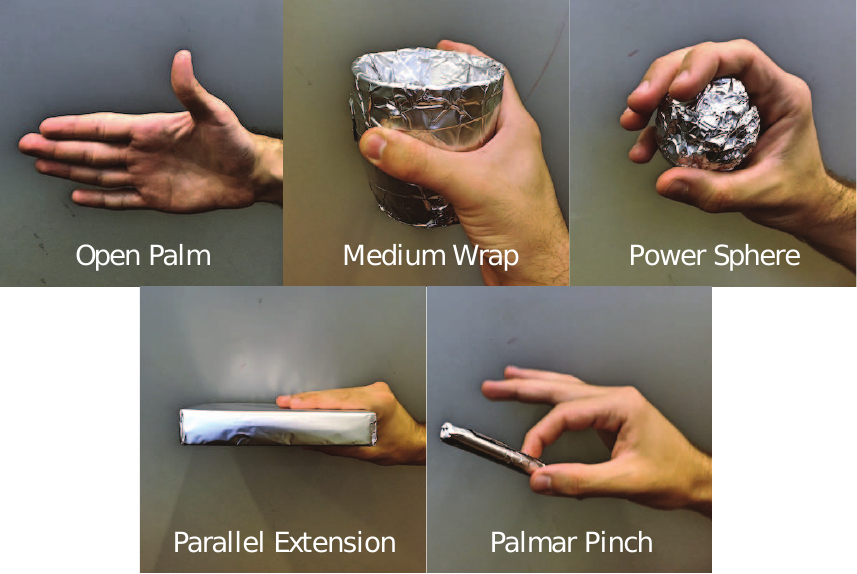}
  \caption{The selected 5 grasp categories.}
  \label{fig:gesture}
\end{figure}

\begin{figure}[t]
  \centering
  \includegraphics[width=3.3in]{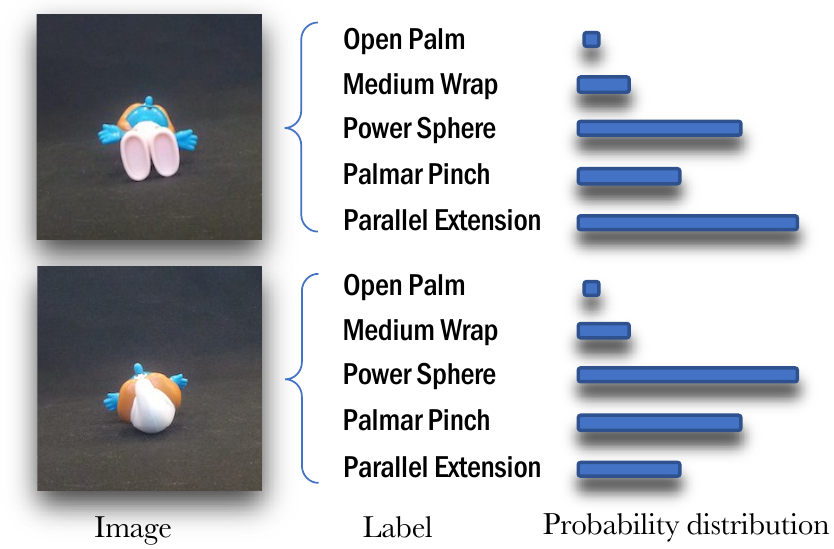}
  \caption{The images and corresponding labels. For each image, there exits a ground truth generated from 11 labelers, which is the probability distribution over 5 grasps. Same object from different views may lead to different distributions.}
  \label{fig:distribution}
\end{figure}

\begin{figure}[t]
  \centering
  \includegraphics[width=3.5in]{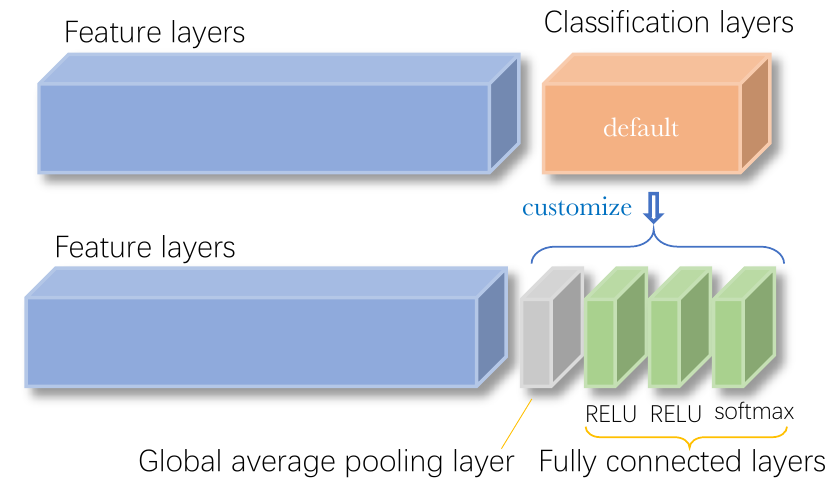}
  \caption{While training CNNs, we freeze all feature layers and replace the default top layers by our customized top layers, and only train those top models added to the fixed features.}
  \label{fig:layer}
\end{figure}

In this section, we provide details on the dataset, architectural specifications and methodology used for transferring the selected ImageNet models to the problem of grasp probability estimation.

\subsection{Dataset}
The data used in this work is based on \cite{RANKCNN}. The dataset consists of 4130 images, which were augmented from 413 hand-perspective images of 102 ordinary objects including office and daily supplies, utensils, and complex-shaped objects. In the process of learning, the environments and image backgrounds could differ from the practice, which may introduce interference and redundant information during the feature extraction. To focus the learning on the object shape instead of the mutative background, the objects were segmented from the raw images which makes the training independent and orthogonal from the random environment. In addition, to enlarge the dataset and add arbitrary background information to respond to the variable environments, the segmented objects were superposed on a series of Gaussian-noised background. The specific augmentation processes are as follows: first the objects were cut out of the raw images and randomly blurred; then, background of Gaussian noise with random variance was added to the bottom of the segmented object to increase the system robustness to different actual backgrounds; finally, the segmented and blurred object was placed in the Gaussian noise background at random location to form the final training image. In addition, the label set is limited within 5 gestures (Open Palm, Medium Wrap, Power Sphere, Parallel Extension and Palmar Pinch) based on their compliance with robotic implements and also their coverage ability for common objects of daily lives due to the similarity with the other grasp types. The \autoref{fig:gesture} shows the 5 grasp types used. 

Note that the inference problem to be solved is not a hard classification, which would predict a single category. Instead, the inference here needs to predict a probability distribution over grasp types which estimates the suitability of grasp types for a given object. This distribution has to match the distribution observed in the ground truth data. Ground truth data was collected by asking each labeler to rank the 5 grasp in decreasing order of preference to grab the object. These labels are obtained from 11 individuals. We used the most relevant grasp type among 5 grasp types for each object to create a probability distribution. The probability of grasp type $i$ where it is chosen by $n$ labelers from total of $N$ labelers is: 
\begin{equation}
p_i = n/N    
\end{equation}
\autoref{fig:distribution} demonstrates examples of the image data with their corresponding probability distribution over 5 possible labels.

\subsection{Details on Transferring and Network Topology}

To create networks suitable for the new domain and task, the features from the original pretrained model are extracted. This means the top layers, also known as the classifier part of the network, are excluded and not transferred. On top of the transferred features, a Global 2-D Average Pooling is added to reduce the spatial dimension followed by 3 Fully Connected layers with 256, 128 and 5 neurons respectively. The FC layers are stacked with ReLU activations except the last one wherein a Softmax is used as it transforms the prediction into a probability distribution. 

Our methodology for training applied to all networks is two folded: (1) Firstly, we freeze all feature layers and replace the default top layers by our own customized top layers, and only train those top models added to the fixed features, as shown in Figure \ref{fig:layer}. These layers are initialized with random wights using Xavier's method \cite{xavier}. Then the optimizer with learning rate of 0.001 trains the network for 50 epochs. (2) After the top FC layers are trained on the target dataset, all other layers are unfreezed and the whole network is trained with a lower 0.0001 learning rate for another 50 epochs.

\subsection{Training Setup}

In both steps of training FC layers and fine-tuning the whole network afterwards, the models were trained using the batch size of 32 images using Adaptive Moment Estimation (Adam) optimizer \cite{adam}. As the format of labels and predictions are both probability distributions over grasp types, in order to minimize the difference between the current prediction and the ground truth we use the cross entropy \eqref{loss} as the loss function, which measures the error between two distributions:
\begin{equation}
\label{loss}
\begin{split}
    loss = -\frac{1}{n}\sum_{i=1}^{n}{\left[ y_i \log{(p_i)} + (1-y_i)log{(1-p_i)}  \right]}
\end{split}
\end{equation}
where $n$ is the number of categories; $y_i$ and $p_i$ are predicted and ground truth probabilities of grasp type $i$, respectively.

\subsection{Evaluation Metric}
The one-hot encoded labels used in conventional hard classifications problems fail to capture the true nature of grasp types since they can only encode one grasp type while gestures capable of grabbing a given object can be more than one. Moreover, absolute zero and one values used for labels in soft-classification would also fail to represent preference of one grasp type over the others. Each potential grasp type for a specific object might not have the same preference to the human over the others. To avoid this information loss, the value for each grasp type is represented as probabilities that sum up to 1. Therefore, there needs to be an evaluation metric that can calculate the error given the prediction and ground truth probability distributions. However, for our probability estimation problem, choosing an evaluation metric becomes challenging since it is not possible to clip probabilities to absolute values since it will result in significant information loss. In result, the evaluation metrics of conventional hard classification, such as Top-1 accuracy, cannot be applied.

To have a simple yet powerful metric, we propose angular similarity for evaluating the effectiveness of the model:

\begin{equation}
    sim(u, v) = ( 1 - 2 \cdot arccos(\frac{u \cdot v}{\lVert u \rVert \lVert v \rVert})/\pi )
\end{equation}
where $u$ and $v$ are vectors of probability distributions for prediction and the ground truth which are all positive and sum to 1. The angular similarity measures the angle between two given vectors which ranges from 0 to 1. A higher similarity indicates that the vectors are closer to each other, i.e. that the ground truth and the estimated distribution match more closely. 

As an example to evaluate how probability values impact our proposed method, given ground truth probability $true=(1, 0, 0, 0, 0)$, $pred=(1,0,0,0,0)$ yields the highest value of 1. For $pred=(0.87,0.13,0,0,0)$ the evaluation metric yields good value of 0.9 and $pred=(0.76,0.24,0,0,0)$, $pred=(0.67,0.34,0,0,0)$, $pred=(0.58,0.42,0,0,0)$ and $pred=(0.5,0.5,0,0,0)$ result in angular similarity of 0.8, 0.7, 0.6 and 0.5 respectively. Note that $pred=(0,1,0,0,0)$ results in the lowest value of 0 which implies importance of the probability values order. It is also noteworthy to provide and example which examines both order and values of the predicted probabilities. Considering $pred=(0.2,0.2,0.2,0.2,0.2)$, will result in low performance of 0.3.

Moreover, comparing angular similarity to cosine similarity as an orientation based metric, it is a function of a proper distance when subtracted from 1, whereas in cosine similarity for small angles the resulting cosine values are very similar.

\section{Results}
\label{sec:results}

\begin{figure}
\centering
\begin{subfigure}[b]{1\textwidth}
   \includegraphics[width=1\linewidth]{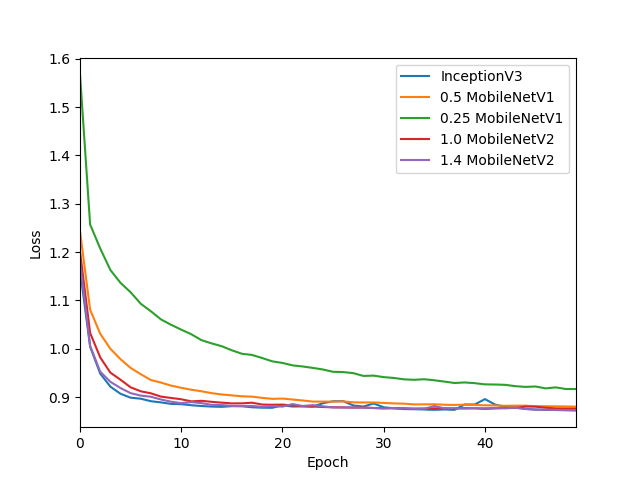}
   \caption{}
   \label{fig:loss} 
\end{subfigure}

\begin{subfigure}[b]{1\textwidth}
   \includegraphics[width=1\linewidth]{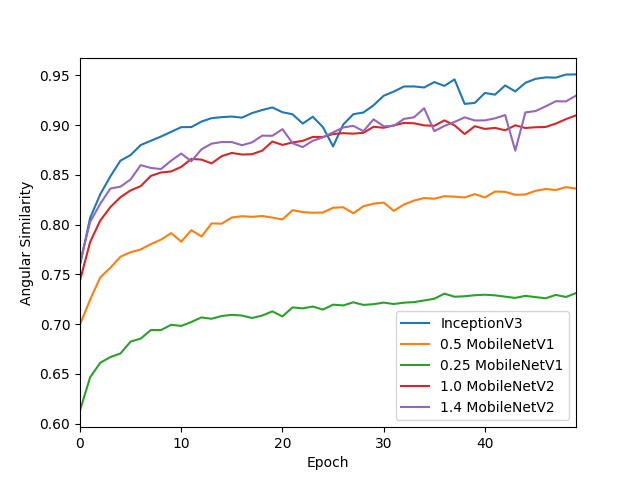}
   \caption{}
   \label{fig:acc}
\end{subfigure}

\caption[Loss and Validation of Pareto Models]{(a) Fine-tuning loss of efficient models. (b) Validation angular similarity of the efficient models}
\end{figure}

\begin{figure}[t]
  \centering
  \includegraphics[width=\linewidth]{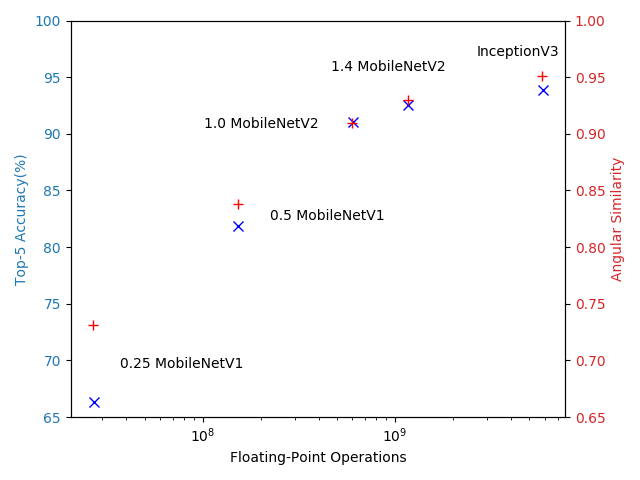}
  \caption{Number of Floating point operations vs. accuracy evaluation of Pareto models before (blue crosses) and after (red pluses) training. Accuracy axis have the same range for a fair comparison.}
  \label{fig:transform}
\end{figure}

We trained the proposed method using TenosrFlow on Pareto efficient models including InceptionV3, MobileNetV1 with 0.25 and 0.5 width multipliers ($\alpha$) and MobileNetV2 with 1.0 and 1.4 width multipliers. Models were trained over 80\% of dataset with batch size of 32 images for 50 epochs and validated on the 20\% rest of the dataset. To monitor how well the training is performed for each model over iterations, the training and validation curves are depicted.

\autoref{fig:loss} shows the fine-tuning cross entropy loss over the number of epochs for all models. The loss curves for all models converge as expected which demonstrate the model is well trained. However, the final cross entropy loss for 0.25 MobileNetV1 is higher than other networks due to significant reduction of number of learnable parameters.

To evaluate how precise each model is, the angular similarity comparison for the Pareto models are provided in \autoref{fig:acc}. The vertical axis is the validation angular similarity for each model over epochs, and we find that the ranking of the trained models with respect to their angular similarity is in total resemblance with the original pretrained models' accuracy on Imagenet.

\autoref{fig:transform} compares the accuracy and performance of the selected models before and after transferring. On the left axis, the Top-5 accuracy of the pretrained model are provided. The right axis also depicts the angular similarity of the selected models after applying transfer learning. The number of floating point operations for each network in the source and target domains were calculated which is observable by the slight shift of the grasp probability estimation networks. This is due to the fact that the imageNet classification layers (fully connected) were replaced with the grasp estimation layers, which contain fewer neurons.

Moreover, since the number of operations related to the top layers of the pretrained models are much fewer than those of the extracted features, and the fact that the same amount of computation is required for the added top layers, the total number of operations does not change significantly when transferring to the new dataset. 

As depicted in \autoref{fig:transform}, there is a trade-off between the computational demand and the accuracy of the Pareto models. In a computationally limited application with low intolerance for latency, networks with lower number of floating point operations such as 0.25 MobileNetV1 are preferred. However, as the models become less computationally intensive, they reach closer to the accuracy of a uniform random generator (0.5 angular similarity). 

To compensate for EMG deficiencies, there needs to be an accurate yet efficient model for the visual classifier. While both InceptionV3 and 1.4 MobileNetV2 provide impressive angular similarity of 0.95 and 0.93 respectively, with 0.02 loss in angular similarity MobileNet has 19.88\% ({\raise.17ex\hbox{$\scriptstyle\sim$}} 1/5) of Inception's total number of floating-point operations, hence is expected to execute faster.

\section{Conclusion}
\label{sec:conclusion}

This work aims to restore the lost ability of lower arm amputees using robotic prosthetic hands via pre-defined grasp types. Our system aims to fuse probable grasp types based on visual information with human intent through EMG measurements. With the focus on the visual classifier, our approach utilizes transfer learning of ImageNet models to predict grasp type distributions given the objects visible from a camera integrated into the robotic hand. 

To have an efficient yet accurate prediction, we studied several state-of-the-art models and excluded inefficient networks as a source of transferring knowledge. We also retrained the efficient networks on probabilistic labels instead of hard/soft ground truth labels to have an accurate representation of grasp types. This way, multiple grasp types with different preferences can be represented, suitable for grasp detection problem. To provide a suitable evaluation for the probability estimation, we proposed angular similarity as an intuitive evaluation metric. We also observed that the relative ordering of the selected models in terms of error/performance stays the same using the proposed metric after transferring. Using the proposed method, we selected the best neural network architecture for our prosthetic hand to enable efficient performance along with the EMG inference data. Using 1.4 MobileNetV2 provides 0.93 angular similarity with 20\% of Inception's total number of floating-point operations, improving the performance by 5.02x with 0.02 loss in angular similarity.

\section{Acknowledgement}
This work is partially supported by NSF (CNS-1544895 at NEU; CNS-1544636 at WPI, CNS-1544815 at HMS).

%
%
%
\bibliographystyle{splncs04}
\bibliography{refs}

\end{document}